\documentclass[10pt,twocolumn,letterpaper]{article}

\usepackage{iccv}
\usepackage{times}
\usepackage{epsfig}
\usepackage{graphicx}
\usepackage{amsmath}
\usepackage{amssymb}
\usepackage{comment}
\usepackage[utf8]{inputenc}

\usepackage[pagebackref=true,breaklinks=true,colorlinks,bookmarks=false]{hyperref}

\iccvfinalcopy 

\def\iccvPaperID{21} 

\ificcvfinal\fi
\begin{document}
	
\title{Robust Semantic Segmentation with Ladder-DenseNet Models}

\author{Ivan Krešo \qquad Marin Or\v{s}i\'{c} \qquad Petra Bevandi\'{c}
  \qquad Sini\v{s}a \v{S}egvi\'{c}\\
  Faculty of Electrical Engineering and Computing\\
  University of Zagreb, Croatia\\
  {\tt\small name.surname@fer.hr}
}

\maketitle

\begin{abstract}
We present semantic segmentation experiments 
with a model capable to perform 
predictions on four benchmark datasets: 
Cityscapes, ScanNet, WildDash and KITTI.
We employ a ladder-style convolutional architecture
featuring a modified DenseNet-169 model
in the downsampling datapath,
and only one convolution in each stage
of the upsampling datapath.
Due to limited computing resources,
we perform the training only on
Cityscapes Fine train+val, ScanNet train, 
WildDash val and KITTI train.
We evaluate the trained model on the 
test subsets of the four benchmarks
in concordance with the guidelines of 
the Robust Vision Challenge ROB 2018.
The performed experiments reveal
several interesting findings 
which we describe and discuss.
\end{abstract}

\section{Introduction}

Semantic image segmentation provides
rich information on surrounding environment,
which presents clear application potential in many domains.
However, there are challenges which are still to be solved
before this exciting technique becomes ready for the real world.

Firstly, assessing the prediction uncertainty is necessary 
if we wish to be able to warn downstream processing elements 
when model predictions are likely to be wrong.
Half of the solution consists in detecting image regions
which are completely different from the training images
and therefore fall in the category of out-of-distribution examples \cite{hendrycks17iclr}.
The other half of the solution is to detect regions 
which are poorly learned or inherently hard to classify 
\cite{kendall17nips}, that is to recognize parts of the scene 
where our models consistently fail to produce correct results.

Furthermore, there is little previous research 
on semantic segmentation models which are suitable 
for recognizing different kinds of environments
in images with no photographer bias.
Before performing experiments presented in this report
we did not know whether such models could be trained
without one domain knowledge interfering with another.
We also did not know how much capacity is required
in order to produce state of the art predictions in different scenarios.

The Robust Vision Challenge provides 
a good testbed to address these questions.
Diversity of the included datasets poses challenges 
to models which may be biased towards a single dataset 
while not generalizing well on others.
Simultaneous training on diverse datasets 
provides an opportunity to learn representations 
which produce good and robust results 
in a multitude of environments.

This report presents main findings gathered
while participating in the ROB 2018 challenge.
We describe the employed model \cite{kreso17cvrsuad}, 
detail the training procedure
and present main insights obtained during our experiments.

\section{Datasets}

We train our common model on the following four training subsets: 
Cityscapes Fine train+val, WildDash, KITTI train and ScanNet train.
Due to limited computing resources and limited time 
we chose to leave other prospective datasets for future work.
Thus, we did not train on Berkeley Deep Drive, 
Vistas and Cityscapes coarse, 
although we did initialize our training 
with parameters learned on ImageNet \cite{russakovsky15ijcv}.
The rest of this section provides a brief overview 
of each of the four training datasets.

\subsection{Cityscapes}

The Cityscapes dataset \cite{cordts15cvpr}
contains images from the driver's perspective 
acquired in cities from Germany and neighbouring countries.
The dataset provides 2MPx images 
split into train, val and test subsets, 
where the semantic labels for the test subset 
are not publicly available.
There are 19 label classes used for evaluation 
which we train upon.
Train and val subsets consist of 2975 and 500 
finely annotated images, respectively.
The dataset also provides 20\,000
coarsely annotated images 
which we do not use in any of our experiments.

\subsection{WildDash}

The WildDash dataset contains a small selection 
of worldwide driving images with a strong potential 
to present difficulties for recognition algorithms.
The dataset contains 70 validation and 156 testing images
which are grouped into ten specific hazardous scenarios 
such as blur, windscreen interference, lens distortion etc.
The image resolution is 1920x1080px while the 
semantic annotations follow the Cityscapes labeling policy.
As in other datasets, the test labels 
are not publicly available.

This dataset is unique since the test subset contains
a number of heavily distorted and out-of-distribution images 
whose correct pixel-level predictions 
may either be the exact class or the class "Void"
(both cases are counted as true positives).
The negative images must be treated 
in the same way as the rest of the dataset,
which suggests that aspiring models 
should include a method for detecting 
out-of-distribution patches in input images.

Due to some labeling inconsistencies 
(e.g.\ terrain vs vegetation and car vs truck)
we follow the ROB practice and evaluate 
WildDash performance with the category iIoU metric.

\subsection{KITTI}

The KITTI dataset \cite{geiger12cvpr} 
has been collected in Karlsruhe, Germany 
while driving through the city itself 
and the surrounding area.
It provides 200 images for training 
and 200 images for testing at 1242x370px.
The dataset uses the Cityscapes labeling policy,
same as the previous three driving datasets.

\subsection{ScanNet}

The ScanNet dataset \cite{dai17cvpr} 
is the only indoor dataset used.
It is by far the largest dataset of the four, 
consisting of nearly 25\,000 training and 962 test images.
This introduces a large distribution disbalance 
between indoor and driving labels 
which needs to be suitably handled.
There are 20 semantic classes common to indoor scenery.
The image resolution varies, while 
most images have 1296$\times$968px.

\section{Method}
\label{sec:method}
We use a custom fully convolutional model 
based on DenseNet-169 \cite{huang17cvpr}.
The model features a ladder-style upsampling path
\cite{valpola14corr,ronneberger15arxiv,lin16arxiv,kreso17cvrsuad} 
which blends high quality semantics of the deep layers
with fine spatial detail of the early layers.
The model produces logits at 4$\times$ subsampled resolution
which we upsample to the input resolution with bilinear interpolation,
and feed to the usual cross-entropy loss 
with respect to one-hot groundtruth labels.

The main differences with respect to 
our previous work \cite{kreso17cvrsuad} are as follows.
First, we have replaced all concatenations in the upsampling path with summations.
This increased the efficiency of our upsampling path 
without losing any IoU accuracy in validation experiments. 
Following that, we have increased the number of convolution filters
in the upsampling path from 128 to 256 because we assumed 
that 128 feature maps could lead to underfitting 
while training on all four ROB 2018 datasets.
Second, we replace the context layer at the end of the downsampling path 
with a spatial pyramid pooling block very similar to \cite{zhao17cvpr}.
Third, we remove the auxiliary loss at the end of the downsampling path 
and replace it with a novel auxiliary loss which we call the pyramid loss. 
The components of the new loss are defined in terms of softmax predictions 
obtained from representations obtained right after 
feature blending units within the upsampling path 
at 64$\times$, 32$\times$, 16$\times$ and 8$\times$ subsampled resolution.
We do not upsample these auxiliary predictions 
to the input resolution as in the main loss.
Instead, we define the pyramid loss components
as cross-entropy between softmax predictions
and the groundtruth distribution over class labels in the N$\times$N boxes 
where N denotes the corresponding subsampling factor.

During training we oversample Cityscapes, KITTI and WildDash images multiple times in
order to achieve 2:1 example ratio with respect to ScanNet in each epoch.
Mixing outdoor and indoor images into each batch was very important 
in order to get batchnorm moving population statistics 
that correctly approximates batch statistics on both tasks.
For data augmentation, we apply random scale resize between 0.5 and 2,
random crop with 768x768 window size and random horizontal flip with 0.5 probability.
These hyperparameter values are shared for all datasets.
We used the Adam optimizer \cite{kingma14corr} with the base learning rate
of $4e^{-4}$ and additionally divide the learning rate by a factor of 4 for
the ImageNet pre-trained subset of parameters. 
The contribution weight of the pyramid auxiliary loss was set to 0.4.
We set the batch size to 8 and train the common model for 200k iterations.
The training took around 3 days on one Titan Xp GPU.

\section{Results}

We apply the common model to the test subsets of all four datasets,
collect the model predictions and map them to the required formats
of the individual benchmarks where necessary (Cityscapes).
We analyze the obtained results 
and present the most interesting findings.

\subsection{Mapping predictions to the dataset formats}

A common model for the ROB 
semantic segmentation challenge 
has to predict at least 39 object classes: 
the 19 driving classes from Cityscapes 
and 20 indoor classes from ScanNet.
The benchmark scripts for 
WildDash, KITI and ScanNet datasets 
automatically map foreign class indices 
to the negative classes 
"Void" (Cityscapes) 
or "Ignore" (ScanNet).
The Cityscapes benchmark is oblivious of ScanNet indices
and therefore we had to manually remap 
ScanNet predictions to the class "Void"
(we had very few such pixels as shown in 
 Table\ \ref{tab:cross_dataset_results}).

Note that the negative classes ("Void" and "Ignore")
are separate form the 39 object classes.
Predictions of negative classes do not contribute
to true positives on Cityscapes, KITTI and ScanNet,
however they still may improve performance 
since they do not count as false positives.
However, negative predictions constitute true positives
in several WildDash images.

\subsection{False-positive detections of foreign classes}

This group of experiments explores
incidence of false negative detections
due to predictions of foreign classes.
This can be easily evaluated on the test datasets
because there is no overlap between 
indoor and driving classes. 
We look at the number of "driving" pixels
in the ScanNet test dataset as well as at 
the number of "indoor" pixels 
in Cityscapes test, WildDash test 
and KITTI test.
The results are summarized in 
Table \ref{tab:cross_dataset_results}.
The results show that, perhaps surprisingly, 
cross-dataset training resulted in negligible 
increase of false positive detections
due to sharing the model across 
different kinds of scenery. 
\begin{table}[htb]
  \footnotesize
  \begin{center}
    \begin{tabular}{c|c|c}
      & driving classes (\%) & indoor classes (\%) \\
      \hline\hline
      Cityscapes & 99.857 & 0.143 \\
      WildDash & 97.649 & 2.351 \\
      KITTI & 100 & 0 \\
      ScanNet & $\approx 0$ & $\approx 100$ \\
      \hline
    \end{tabular}
  \end{center}
  \caption{
    Incidence of foreign pixels
    in the test subsets of the four datasets.
    The rows correspond to the four datasets
    while the columns correspond to 
    the two groups of classes. 
    We see that cross-dataset training 
    causes very few false positive pixels
    and therefore results
    in a negligible performance hit.}
  \label{tab:cross_dataset_results}
\end{table}

\begin{figure}[htb]
  \centering
  \includegraphics[width=0.8\columnwidth]{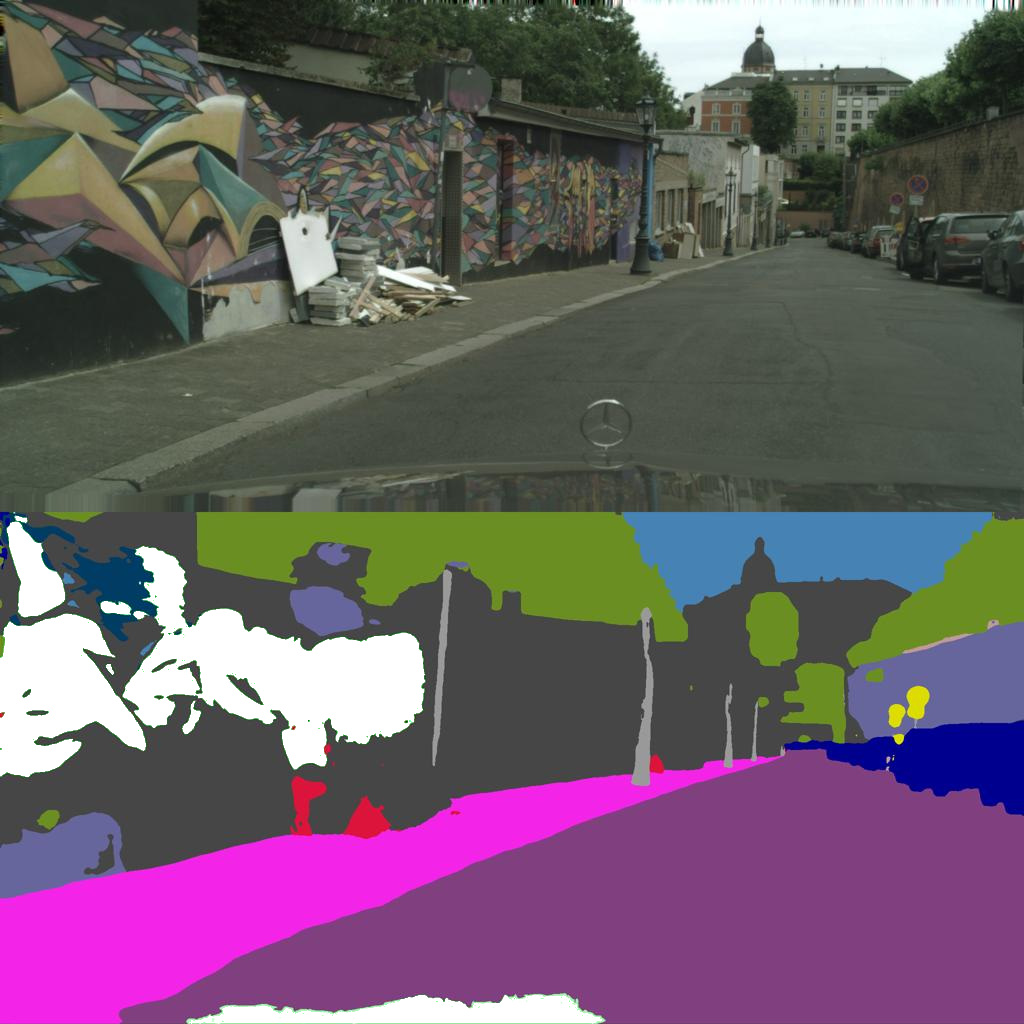}
  \caption{The only Cityscapes test image in which
    a large group of pixels was misclassified 
    into indoor classes.
    The graffiti on the building (white pixels)
    were classified as the indoor class "wall".
  }
  \label{fig:cityscapes_scannet}
\end{figure}

Most foreign pixels in Cityscapes test images
are located on the car hood 
which is ignored during training.
Figure~\ref{fig:cityscapes_scannet} shows 
the only Cityscapes test image with
a relatively large group of predictions
to foreign classes.
There were zero detections of foreign classes on KITTI, 
and only 8 detections of foreign classes on ScanNet.
Most of foreign pixels on WildDash test
are located in negative images
and are therefore treated as true positives
(we explore this in more detail later).

\subsection{Detecting negative WildDash pixels}

\begin{figure}[htb]
  \centering
  \includegraphics[width=0.32\columnwidth]{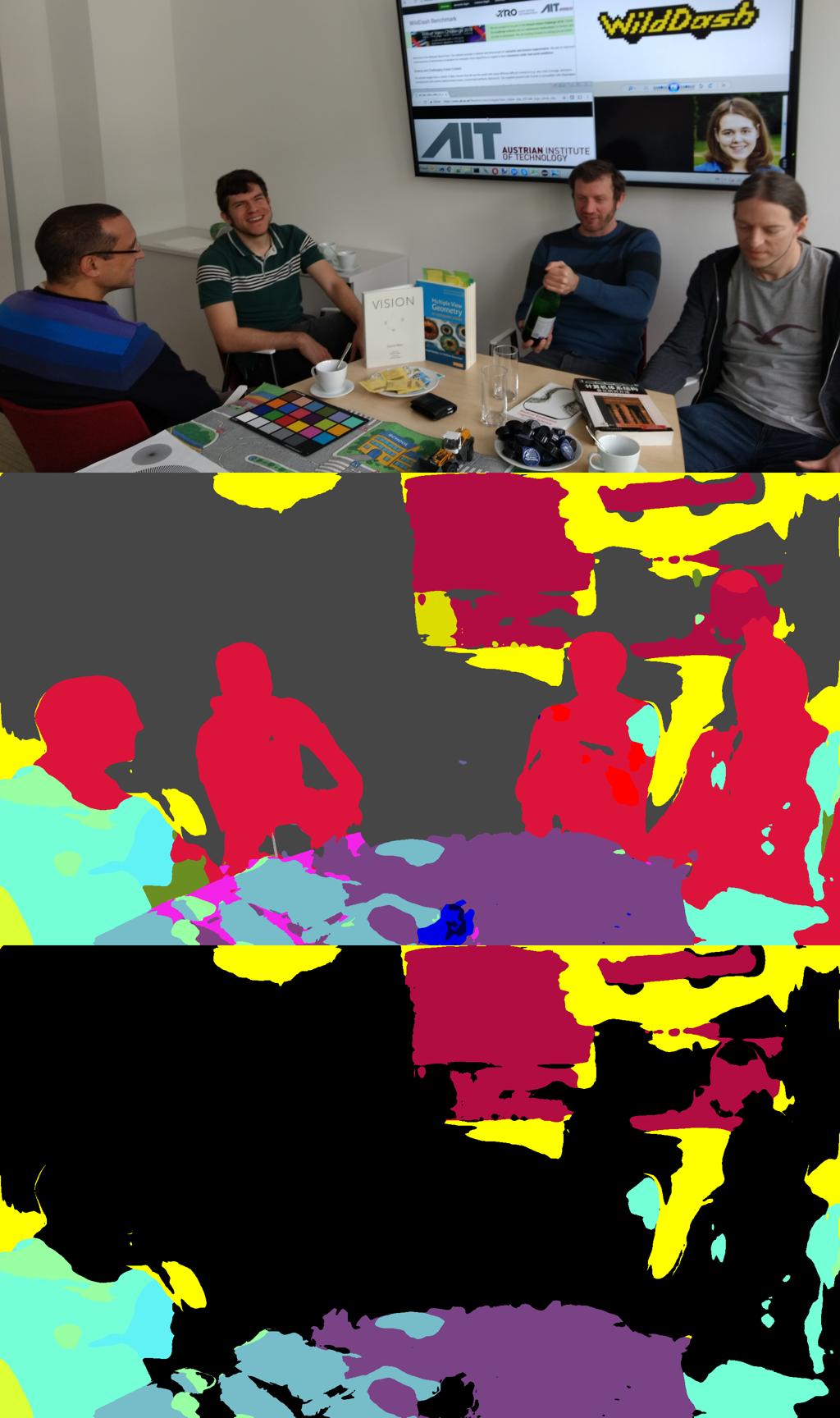}
  \includegraphics[width=0.32\columnwidth]{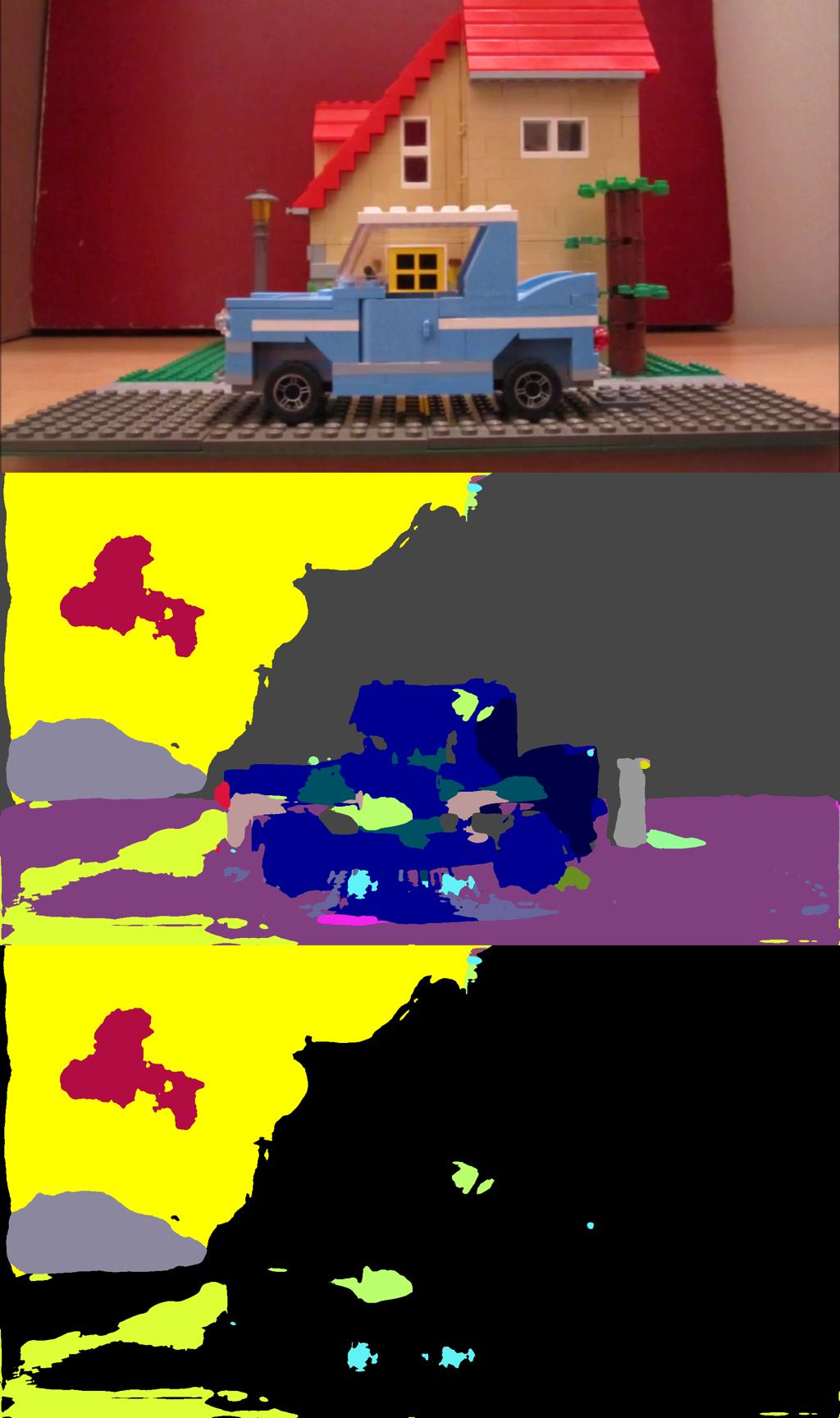}
  \includegraphics[width=0.32\columnwidth]{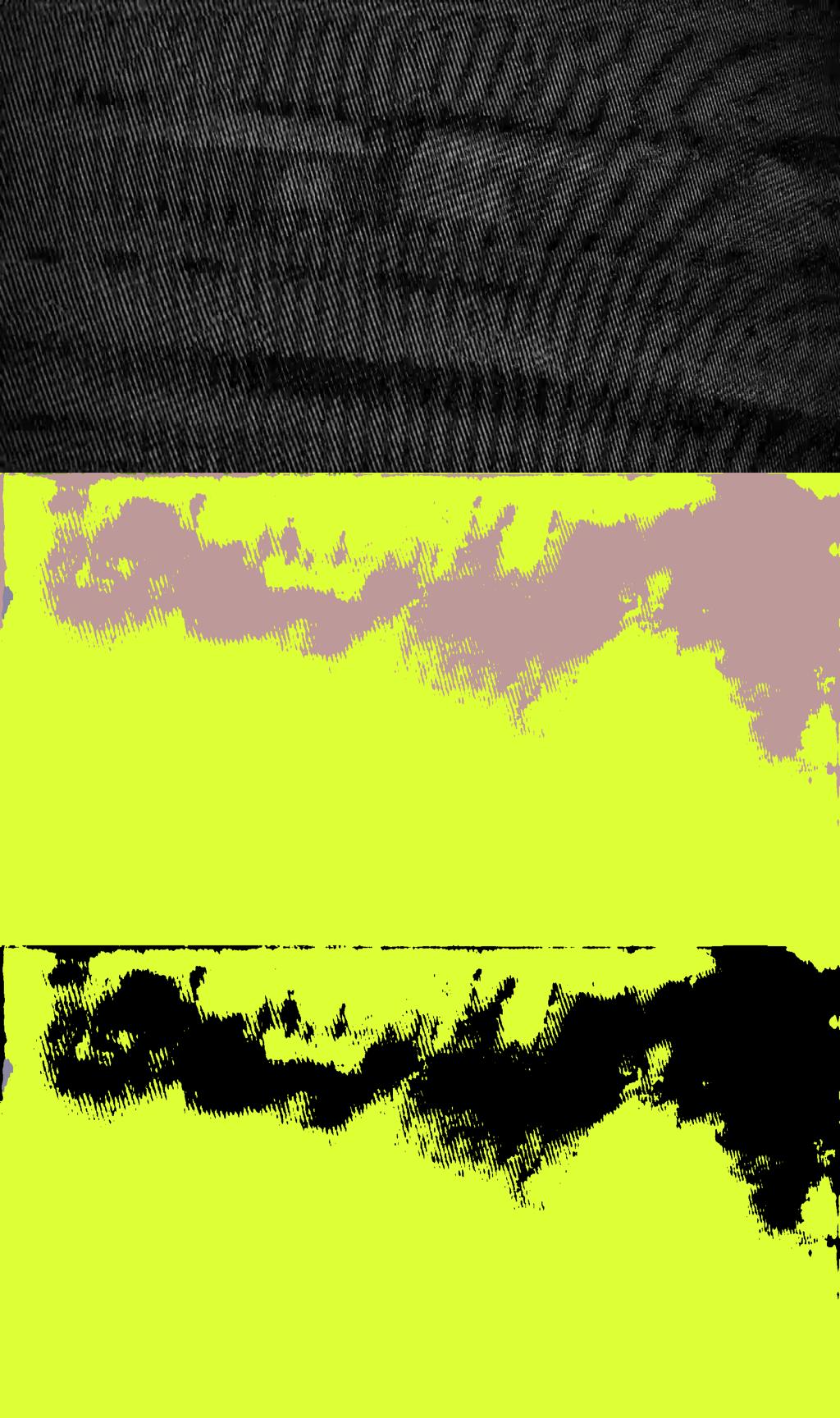}
  \caption{
    Negative WildDash images (top),
    model predictions (middle)
    and out-of-distribution pixels 
    for the WildDash dataset (bottom). 
    In the bottom row we colored all pixels 
    classified as Cityscapes classes in black.
    The remaining colored pixels are treated 
    as the "Void" class during evaluation.}
  \label{fig:wilddash_ood}
\end{figure}

Closer inspection of WildDash test images
revealed that almost all pixels classified as ScanNet 
occur in the negative WildDash images. 
We illustrate three such images 
in Figure~\ref{fig:wilddash_ood}.
The figure shows that Cityscapes detections
are often correct (people, building)
or almost correct (indoor walls as building,
lego pavement as road).

Table \ref{tab:ood_results} shows the difference 
in category iIoU performance 
between our submissions M\_DN and LDN2\_ROB
to the WildDash benchmark.
Both submissions correspond to instances 
of the model described in Section \ref{sec:method}
trained with similar optimization settings.
The submission M\_DN maps all indoor predictions 
to the class Cityscapes "Wall".
The submission LDN2\_ROB leaves 
outdoor predictions as they were,
which means that the benchmark script
automatically maps them to class "Void".
By treating ScanNet predictions 
as the WildDash negative class, 
LDN2\_ROB submission achieved
$9.1$ percentage points improvement in category iIoU.
This improvement gives hope that we could estimate
prediction uncertainty by simply assessing
the likelihood of the foreign classes. 
In other words, we could train our future models
in supervised or semi-supervised manner 
on diverse datasets and use 
the prediction of foreign classes
(that is, classes that are not supposed 
 to appear in this particular image)
as a flag that the predictions are uncertain. 

\begin{table}[htb]
  \footnotesize
  \begin{center}
    \begin{tabular}{l|cc}
      submission ID & ScanNet classes& 
      negative iIoU \\
                    & mapped to& 
      category (\%) \\
      \hline\hline
      M\_DN     & "Wall" & 32.2	\\
      LDN2\_ROB & "Void" & 42.8 \\
      \hline
    \end{tabular}
  \end{center}
  \caption{Results of our two submissions 
    to the WildDash benchmark 
    (M\_DN and LDN2\_ROB).
    LDN2\_ROB improves the performance 
    on negative WildDash test images
    by mapping ScanNet predictions 
    to out-of-distribution pixels.}
  \label{tab:ood_results}
\end{table}

\subsection{Reduction of overfitting in the upsampling path}

Early experiments showed very poor accuracy
of the Ladder-DenseNet architecture 
on the WildDash test dataset.
Further experiments with a simpler model
based on bilinear upsampling 
resulted in better performance.
Consequently, we hypothesized that 
the model with the ladder-style upsampling
suffers from overfitting in the upsampling path.
We attempt to alleviate this problem
by regularizing the model with the pyramid loss
described in Section \ref{sec:method}
(i.e.\ by adding a classification head 
 at each upsampling level),
which resulted in significant improvement.
We illustrate these effects in Table \ref{tab:pyramid_loss}
which shows that the recognition accuracy 
significantly increases when we add pyramid loss.
The table also shows that the benefits reproduce 
on the Berkeley Deep Drive dataset 
(note that we do not train on Berkeley Deep Drive 
 in any of the experiments).
\begin{table}[htb]
  \footnotesize
  \begin{center}
    \begin{tabular}{c|c|c||c|c}
      Dataset & \multicolumn{2}{c||}{WildDash} & \multicolumn{2}{c}{BDD} \\
      \hline
      Pyramid loss & No & Yes & No & Yes \\
      \hline\hline
      Flat & 67.2 & 66.5 & 70.9 & 74.6 \\
      Construction & 18.1 & 16.8 & 51.9 & 52.4 \\
      Object & 13.4 & 24.1 & 29.8 & 34.3 \\
      Nature & 72.1 & 71.9 & 65.5 & 65.6 \\
      Sky & 67.7 & 66.6 & 66.9 & 67.8 \\
      Human & 30.4 & 36.0 & 45.4 & 46.2 \\
      Vehicle & 44.1 & 54.5 & 75.6 & 76.3 \\
      \hline
      mIoU & 44.7 & 48.1 & 58.0 & 59.6 \\
      \hline
    \end{tabular}
  \end{center}
  \caption{Category IoU on WildDash val 
    and Berkeley Deep Drive val
    for the model trained 
    on Cityscapes only.
    We observe large improvements 
    on objects, humans and vehicles.
    Both training and prediction was performed
    on half image resolution in this experiment.
    }
  \label{tab:pyramid_loss}
\end{table}

\begin{figure}[htb]
  \centering
  \includegraphics[width=\columnwidth]{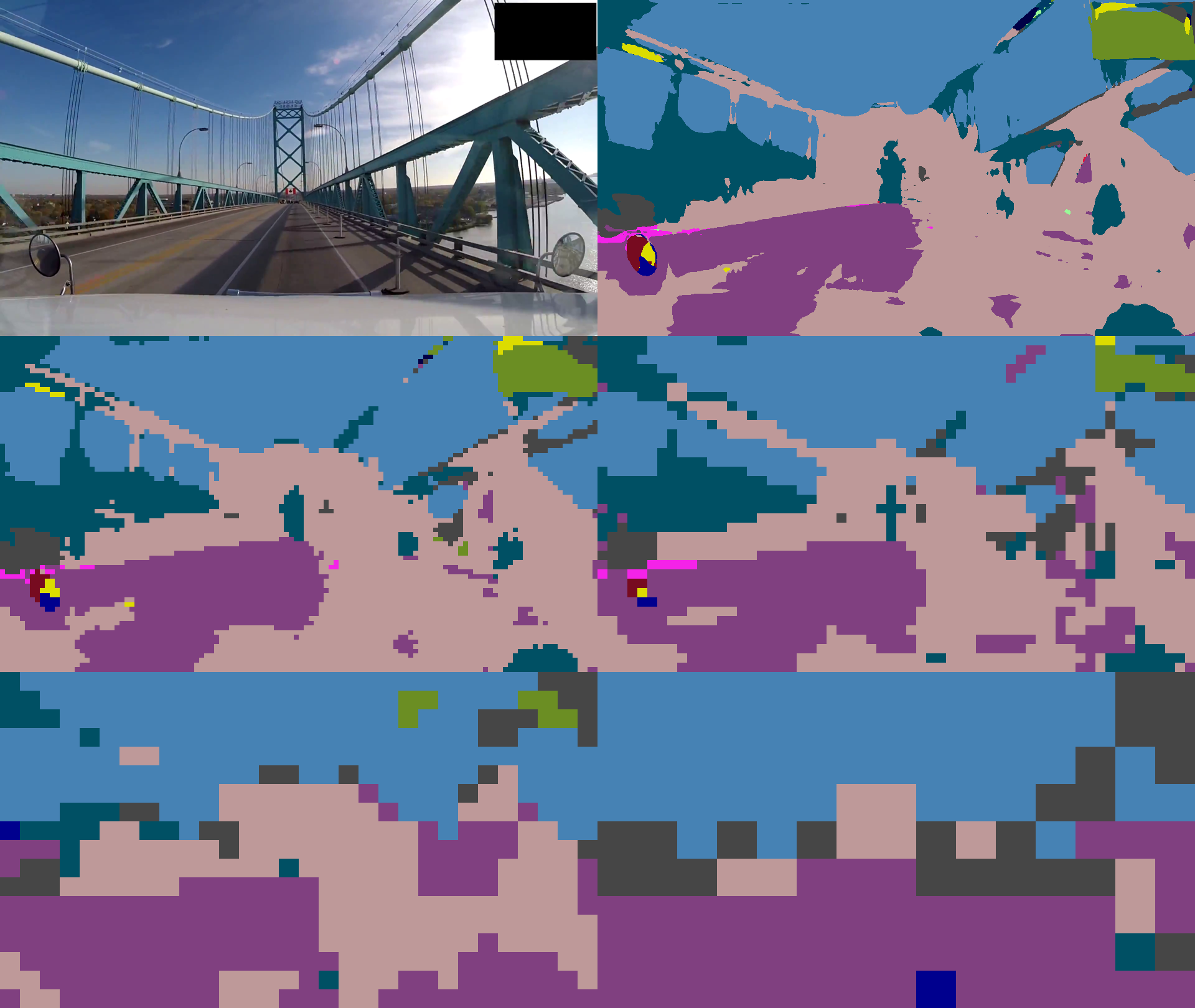}
  \caption{Semantic segmentation predictions 
   of a WildDash val image (top left)
   at different levels of the upsampling path 
   by a model trained on Cityscapes only.
   The resolution increases from bottom to top
   and from right to left.
   We note that the accuracy gets worse
   as we transition from 
   the coarsest resolution (bottom right)
   to the finest resolution (top right).
   Both training and prediction was performed
   on half image resolution in this experiment.
  }
  \label{fig:wilddash_pyramid_loss}
\end{figure}

Further inspection of semantic predictions
along the upsampling path showed that 
some overfitting in the ladder-upsampling remains
despite the pyramid loss.
We illustrate these effects in 
Figure \ref{fig:wilddash_pyramid_loss}.
The image clearly shows that the prediction accuracy
gradually decreases as we transition
towards finer resolutions (top right).
We believe that solving this issue 
might be an interesting direction for future work.

\begin{figure}[htb]
  \centering
  \includegraphics[width=0.32\columnwidth]{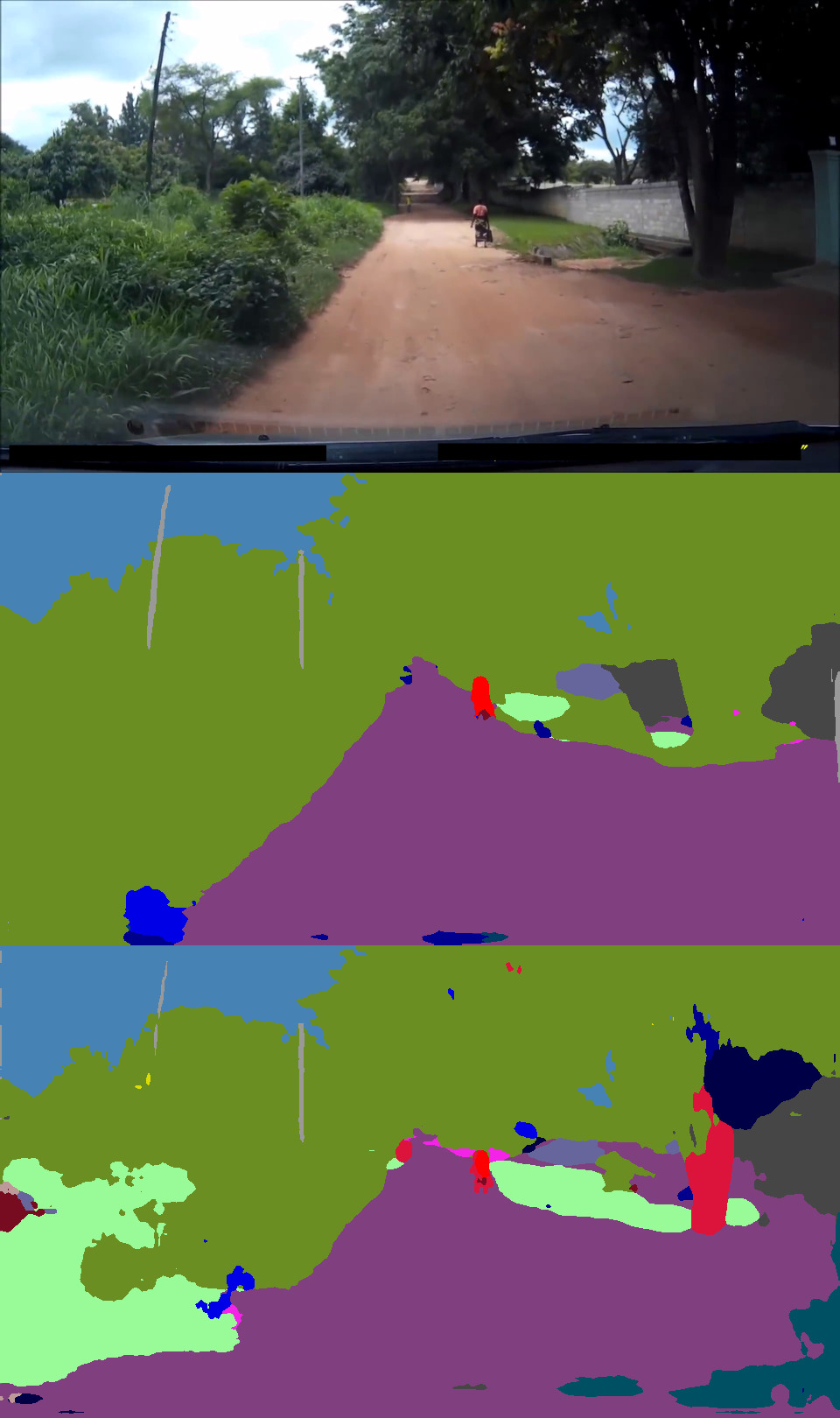}
  \includegraphics[width=0.32\columnwidth]{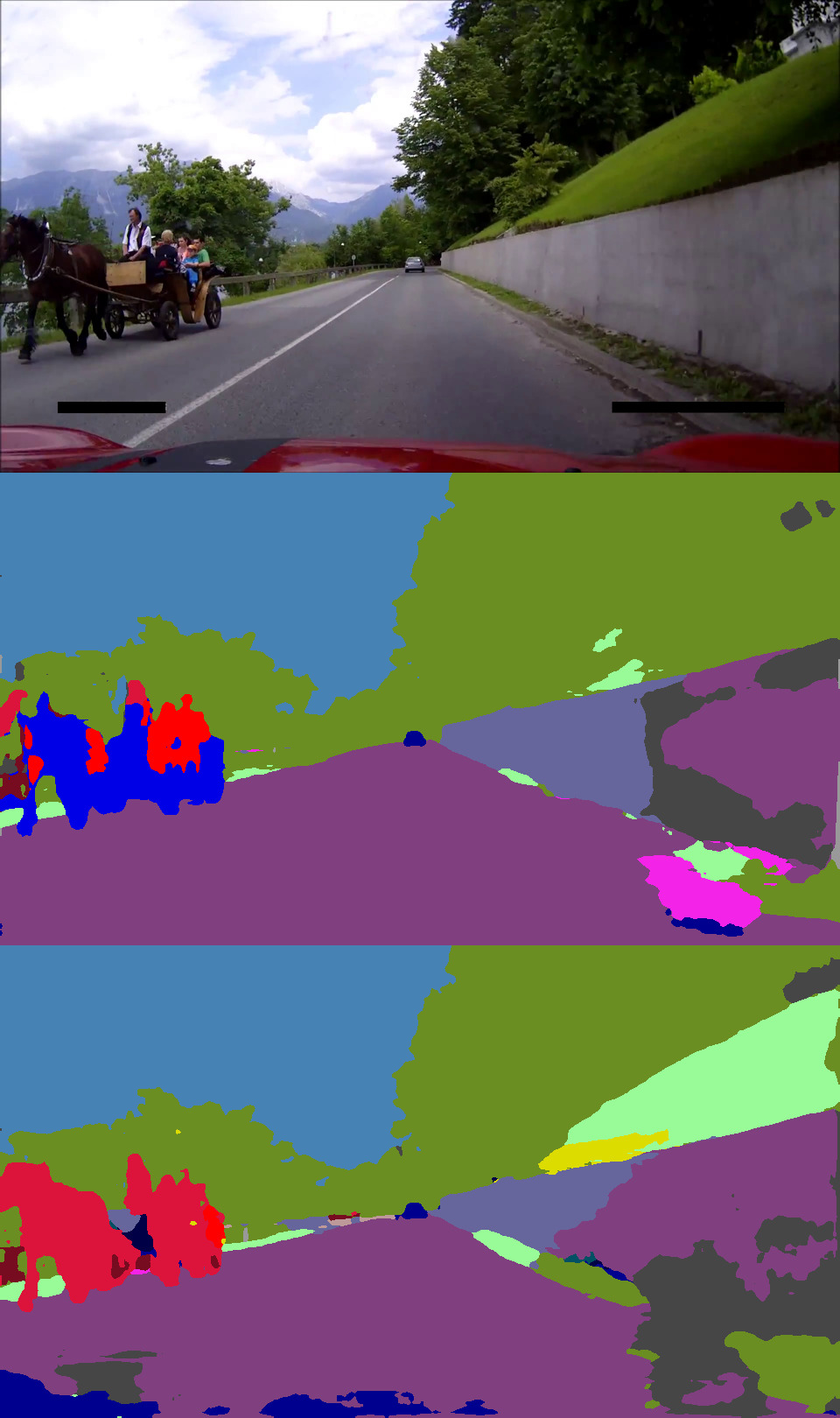}
  \includegraphics[width=0.32\columnwidth]{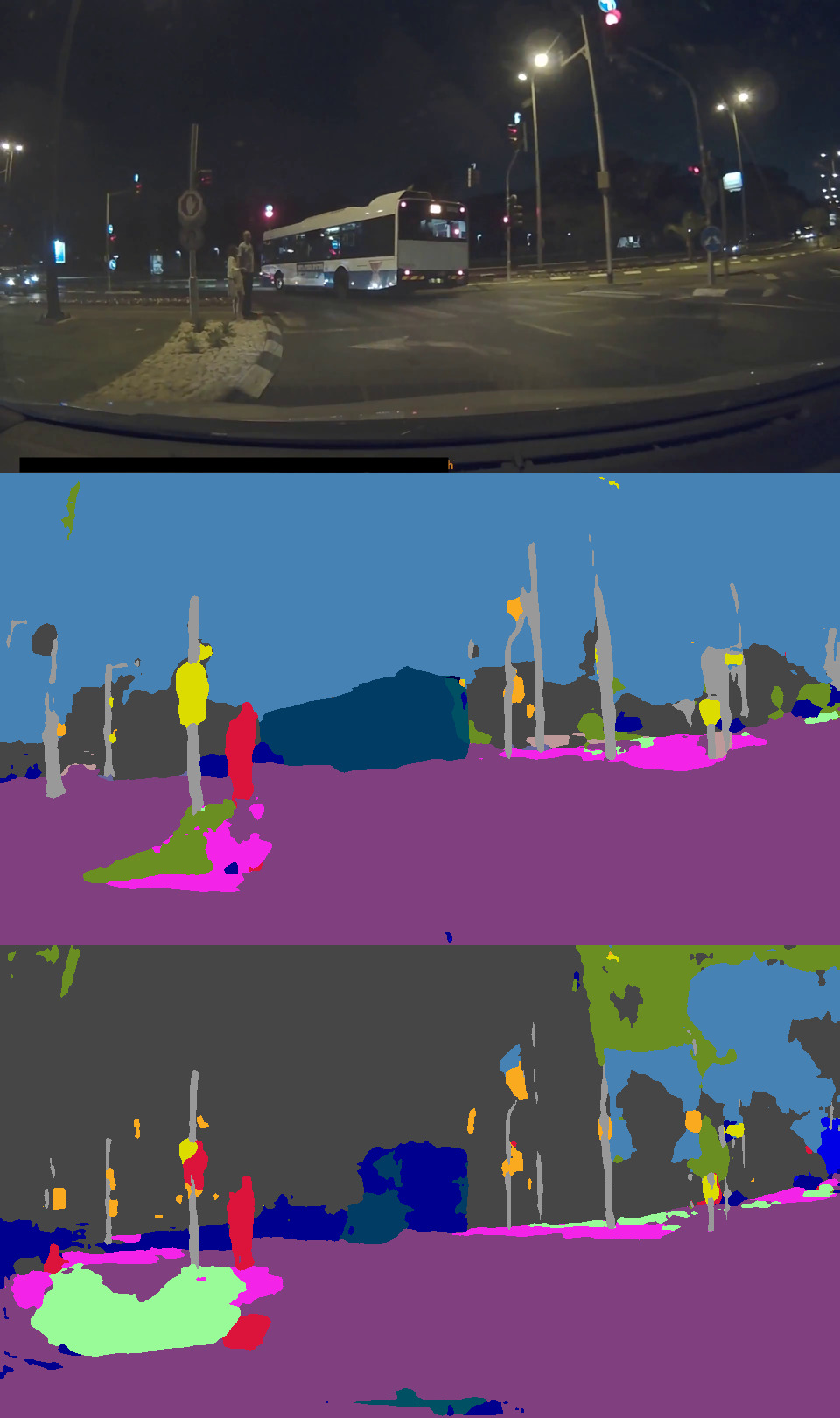}
  \caption{Segmentation results when training on 
    Cityscapes + WildDash val (middle row) 
    vs. training on Cityscapes only (bottom row).
    Both training and predictions was performed
    on half image resolution in this experiment.
  }
  \label{fig:wilddash_val_noval}
\end{figure}

Finally, we show what happens on WildDash test
when we include WildDash val to the training set.
The effect is not easy to quantify 
since WildDash benchmarks allows 
only three submissions per researcher.
We therefore perform qualitative analysis
in several WildDash test images 
and show the results in 
Figure \ref{fig:wilddash_val_noval}.
We see that only 70 WildDash val images
succeeds to significantly impact the model 
despite being used along 3500 images
from the Cityscapes dataset.

\subsection{Overall results}

We have submitted the results 
of our common model 
to the ROB 2018 challenge
under the identifier LDN2\_ROB.
The obtained results on all four datasets
are summarized in Table \ref{tab:rob_results}.

\begin{table}[htb]
  \footnotesize
  \begin{center}
    \begin{tabular}{l|cccc}
      dataset    & metric    & our result & best result & our rank \\
      \hline\hline
      KITTI      & class IoU & 63.5       & 69.6 & 3 \\
      ScanNet    & class IoU & 44.0       & 48.0 & 2 \\
      Cityscapes & class IoU & 77.1       & 80.2 & 2 \\
      WildDash   & category IoU & 54.5    & 59.1 & 3 \\
      \hline
    \end{tabular}
  \end{center}
  \caption{Results of our common model LDN2\_ROB
    at the four semantic segmentation benchmarks.}
  \label{tab:rob_results}
\end{table}

\section{Discussion}

The presented experiments resulted 
in several interesting findings.
Initial experiments with ladder-style models 
resulted in very poor cross-dataset performance.
Closer inspection revealed that small errors
had been multiplying along the upsampling datapath.
This likely occured due to blending convolutions 
being overfit to the Cityscapes urban scenes 
with ideal weather conditions 
and a high-quality HDR camera.
These effects might be even larger 
in models with more capacity
in the upsampling datapaths.
Experiments have showed that this problem
can be successfully mitigated 
with suitable auxiliary losses
and training on the WildDash val subset.

The second interesting result is that ScanNet training
significantly improved recognition of
out-of-distribution pixels on WildDash test.
In fact, many such pixels were detected 
as some of the ScanNet classes
and were therefore treated as true positive predictions.
This raises hopes that future models will be able
to detect unusual parts of the scene for free, 
only by virtue of being trained 
on a more diverse set of classes.

Further, we found that 
simultaneous training on multiple datasets
resulted in virtually no performance hit
with respect to training only on one dataset.
In fact, less than 0.01 percent 
of valid in-distribution pixels in all
three driving dataset test sets were recognized 
as one of the ScanNet indoor classes.
Conversely, only 8 out of 
around billion pixels in ScanNet test
were recognized as one of the Cityscapes driving classes.

Finally, we found that batch composition represents
an important ingredient of cross-dataset training.
The training convergence improved substantially
when we switched from training on single-dataset batches
to training on cross-dataset batches.
We hypothesize that the improvement occurred 
due to more stable training of batchnorm layers.

\section*{Acknowledgements}

This work has been supported by the European Regional Development Fund under the project "System for increased driving safety in public urban rail traffic (SafeTRAM)".

The Titan Xp used in experiments was donated by NVIDIA Corporation.

\newpage

{\small
\bibliographystyle{ieee}
\bibliography{paper}
}


\end{document}